\pdfoutput=1

\documentclass[11pt]{article}
\usepackage{booktabs}

\usepackage[]{acl}
\usepackage{comment}
\usepackage{siunitx}
\usepackage{amsmath}
\usepackage{enumitem}
\usepackage{amssymb}

\usepackage{times}
\usepackage{latexsym}
\usepackage{comment}
\usepackage{xspace}
\usepackage[T1]{fontenc}

\usepackage[utf8]{inputenc}

\usepackage{microtype}
\usepackage{graphicx}
\usepackage{xcolor}

\newcommand{\method}{\texttt{SynRL}\xspace}

\newcommand{\mycomment}[1]{}
\newcommand{\bmE}{\mathbb{E}}

\title{\method: Aligning Synthetic Clinical Trial Data with Human-preferred Clinical Endpoints Using Reinforcement Learning}

\author{
  Trisha Das,
  Zifeng Wang,
  Afrah Shafquat,
  Mandis Beigi,\\
  Jason Mezey,
  Jacob Aptekar,
  Jimeng Sun\\
  \texttt{trishad2@illinois.edu},
  \texttt{zifengw2@illinois.edu},
  \texttt{afrah.shafquat@3ds.com},\\
  \texttt{mandis.beigi@3ds.com},
  \texttt{jgm45@cornell.edu},\\
  \texttt{Jacob.APTEKAR@3ds.com},
  \texttt{jimeng@illinois.edu}
}

\begin{document}
\maketitle

\section{Abstract}

Each year, hundreds of clinical trials are conducted to evaluate new medical interventions, but sharing patient records from these trials with other institutions can be challenging due to privacy concerns and federal regulations. To help mitigate privacy concerns, researchers have proposed methods for generating synthetic patient data. However, existing approaches for generating synthetic clinical trial data disregard the usage requirements of these data, including maintaining specific properties of clinical outcomes, and only use \textit{post hoc} assessments that are not coupled with the data generation process. In this paper, we propose \method which leverages reinforcement learning to improve the performance of patient data generators by customizing the generated data to meet the user-specified requirements for synthetic data outcomes and endpoints. Our method includes a data value critic function to evaluate the quality of the generated data and uses reinforcement learning to align the data generator with the users' needs based on the critic's feedback. We performed experiments on four clinical trial datasets and demonstrated the advantages of \method in improving the quality of the generated synthetic data while keeping the privacy risks low. We also show that \method can be utilized as a general framework that can customize data generation of multiple types of synthetic data generators. Our code is available at \url{https://anonymous.4open.science/r/SynRL-DB0F/}.

\section{Introduction}

Clinical trials aim to assess the impact, effects, and value of medical interventions in patients, often enrolling tens to a few thousand participants, which takes several years to complete \cite{NIH}. While the data generated from trials has high value for downstream analyses aimed at drug safety improvement, bias evaluation, and other meta-analyses \cite{wong2014examination, azizi2021can}, sharing patient data can be difficult due to strict and expensive de-identification and anonymization requirements \cite{NIH}. Generating realistic synthetic patient data offers an opportunity to support in-depth data analysis \cite{choi2017generating}, while overcoming the privacy concerns of trial data sharing. Moreover, it has been demonstrated that augmenting real patient data with synthetic data can improve downstream predictive tasks such as patient outcome predictions \cite{wang2022promptehr}. 

Despite advances in synthetic data generation (SDG) for tabular data \cite{xu2019modeling} and Electronic Healthcare Records (EHR) \cite{wang2022promptehr} that could be applied to generate synthetic clinical trial data, several important challenges remain when applying these to trial data:

\begin{itemize}[leftmargin=*]
    \item \textbf{Targeted data quality audit}: The quality of the data used for training machine learning (ML) models can outweigh the quality of the model and the algorithm being used \cite{wang2020less,zha2023data}. Current practice relies on manually evaluating the generated synthetic clinical trial dataset through a series of \textit{post hoc} estimations of fidelity and utility \cite{beigi2022synthetic}. These current models lack the ability to actively rank synthetic data and associated generation models with human preference.

    \item \textbf{Absence of customized synthesis}: Users of synthetic clinical trial data may have preferred clinical endpoints or outcomes of interest that may be critical to their investigation. For example, a user may be interested in death prediction or adverse event frequency when making use of the synthetic clinical trial dataset. For this scenario, it is necessary that the user receive a customized synthetic dataset that maintains the properties of death-related and adverse event features for downstream tasks. Previous methods have not explored the customized synthesis of clinical trial data when requiring specific properties for clinical endpoints and outcomes.  
\end{itemize}

In this paper, we introduce \method, a new approach to align synthetic data generation models (SDG) with human preferences through reinforcement learning. \method improves the quality of synthetic data generated by a base generator model. Two prominent examples of such base generators are \texttt{TVAE} and \texttt{CTGAN} \cite{xu2019modeling}, which leverage variational autoencoder and generative adversarial network (GAN) principles, respectively, to produce synthetic tabular data. Through a feedback loop, \method adjusts these base generators to align with users' preferred end goals, enhancing the utility of generated data for downstream tasks of interest. This is enabled by the following technical contributions:
\begin{itemize}[leftmargin=*]
\item \textbf{Data value critic function/Reward function} audits the quality of generated intermediate synthetic patient data based on user-specified preferences, i.e., the utility of the generated data when applied to downstream tasks of interest.
 
\item \textbf{Reinforcement learning from feedback} provided by the data critic function obtained in the prior stage that aligns the base generator with the preferences specified by the data critic function.

\end{itemize}

\noindent More broadly, \textbf{\method} can be considered a general framework applicable to a broad range of base generator models, where the aim is to enhance the model's performance within the context of targeted downstream tasks. To avoid confusion, we use \method as the synonym of \method-\texttt{TVAE} throughout the paper which uses \texttt{TVAE} as the base generator. We focus on \method-\texttt{TVAE} because of the interest in using \texttt{VAE} to generate synthetic clinical trial data \cite{das2023twin, beigi2022synthetic} and also show results for \method-\texttt{CTGAN} which uses \texttt{CTGAN} as the base generator to show the flexibility of our framework.

In Section \S \ref{background}, we discuss related papers in the literature. 
We elaborate on the technical details of \method in Section \S \ref{methods}. In Section \S \ref{sec:experiment}, we present the experiment results on four clinical trial datasets: Melanoma, Breast Cancer, NSCLC and CAR-T datasets. The results show that \method improves the utility of synthetic data without compromising fidelity and privacy (Section \S\ref{sec: discussion}). We end with addressing limitations (Section \S \ref{sec: limitation}) and concluding our findings (Section \S \ref{sec: conclusion}).

\subsection*{Generalizable Insights about Machine Learning in the Context of Healthcare}

High quality clinical trial patient data are essential to understanding patient journey and performing valuable analysis including trial protocol optimization \cite{kavalci2023improving} and adverse event prediction \cite{tsarapatsani2023predicting, tong2019machine}. Synthetic patient data presents an opportunity to replace valuable real clinical data (that remain restricted due to privacy concerns) and provide similar insights while preserving patient privacy \cite{doshi2016data}. However, the ability to reproduce downstream analysis from these synthetic clinical trial datasets may vary depending on the synthetic data generation algorithm used \cite{beigi2022synthetic}. Our proposed framework \method presents a solution to improve the quality of the synthetic data generated by fine-tuning using RL to preserve the targeted downstream prediction task of interest. Though the focus for this paper is to address synthetic data generation for clinical trial datasets with sample size limitations, our approach is applicable to any tabular data across healthcare and other domains. Moreover, \method as framework can be generalized to improve performance of any synthetic data algorithm of choice (Section \S \ref{sec:general}).

\section{Related Work}\label{background}

\subsection*{Synthetic patient data generation}
Generating synthetic patient records has emerged as a promising solution to address privacy concerns for sharing personal private data across institutions \cite{wang2022artificial}. The primary focus of research in this field is to use generative adversarial networks (GANs) \cite{choi2017generating,guan2018generation,baowaly2019synthesizing,zhang2021synteg}, Variational Autoencoders (VAEs) \cite{biswal2020eva}, language models \cite{wang2022promptehr}, and diffusion models \cite{yuan2023ehrdiff} to generate synthetic electronic health records (EHRs) that resemble the real EHRs. Building on the success of synthetic EHR generation, recent research has also begun to explore synthetic clinical trial data generation, including tabular clinical trial data \cite{emam2021optimizing,beigi2022synthetic},  longitudinal trial data  \cite{bertolini2020modeling,walsh2020generating}, and for personalized digital twins \cite{das2023twin}. Nevertheless, none of these methods allow users to actively tailor the generation of synthetic data with specific outcome and endpoint properties that will enhance the value of targeted applications.

\begin{figure}[h]
  \centering
  \includegraphics[width=0.45\textwidth]{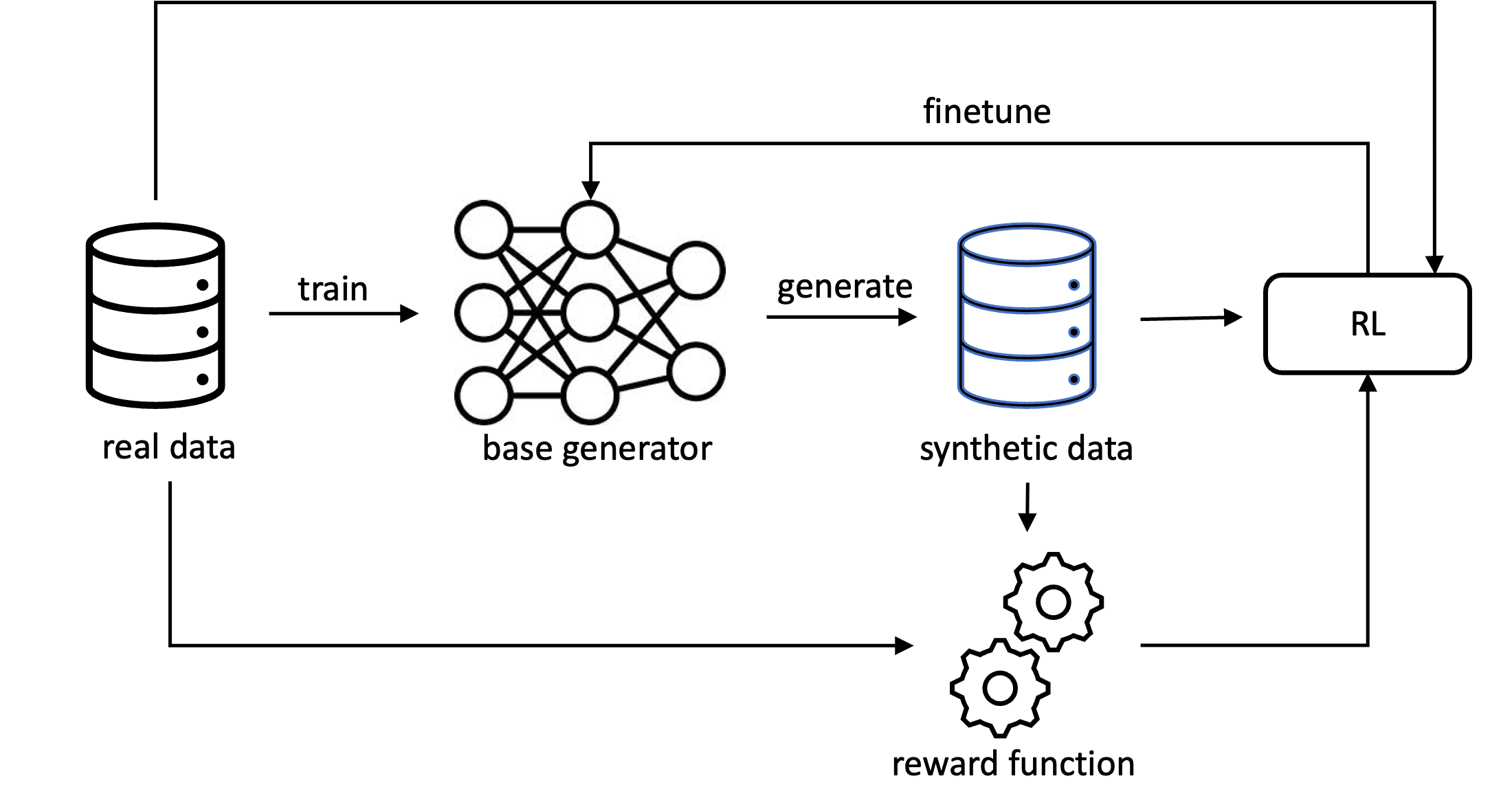}
  \caption{Architecture of \method: it encompasses three main components: the \textit{generator model}, \textit{reward function}, and \textit{RL model}. The generator is first fitted on the real patient record. It then generates the synthetic data to be evaluated and ranked by the reward function. After, the ranking will be the feedback that updates the generator using the RL model.}
  \label{Fig:1}
\end{figure}

\noindent

\subsection*{Reinforcement learning from human feedback}
Reinforcement Learning from Human Feedback (RLHF) is an effective approach for aligning large language models with human preferences, as demonstrated by the success of ChatGPT \cite{ouyang2022training}. Early works proposed the use of reinforcement learning to model human preferences \cite{christiano2017deep,ziegler2019fine} and imitation learning for training language models for long-form question answering \cite{nakano2021webgpt}. Recent advancements in this field are in improving reinforcement learning algorithms over Proximal Policy Optimization (PPO) \cite{schulman2017proximal,dong2023raft}, simplifying the aligning process by discarding reinforcement learning  \cite{yuan2023rrhf} or exploring the mathematical foundations of RLHF \cite{go2023aligning,zhu2023principled}. In some other works, RLHF has been extended to visual generative models \cite{hao2022optimizing,lee2023aligning,wu2023better}. While in text generation, the common practice for RLHF is to crowdsource text labeling tasks, using this approach to label and rank synthetic patient clinical trial data is much more challenging due to the vague definition of synthetic patient record quality and the requirement for domain knowledge. Although a recent study \cite{ganguli2023capacity} has shown the possibility of automatically gaining feedback from AI algorithms to improve language models, aligning synthetic clinical trial data generation with human preferences in an automatic way remains unexplored.

\section{Methods} \label{methods}

\subsection*{The \method framework}\label{sec:synrl_framework}
The purpose of the proposed method is to improve the quality of synthetic clinical trial data for applications requiring specific outcome and endpoint properties by aligning SDG models with the user-specified objectives. Our method allows ranking of the generated synthetic data and provides feedback to the base generator model. Specifically, our model evaluates the synthetic records according to the utility value of important clinical endpoints (e.g., death, specific adverse events, etc.) and tries to improve the synthetic data generator based on the utility scores and other specified properties of the synthetic data using reinforcement learning.  It encompasses the following steps:
\begin{itemize}[leftmargin=*]
\item \textbf{Step 1:} A synthetic data generator model (e.g., TVAE, CTGAN) generates synthetic clinical trial data. We call this the \textbf{generator/base generator}.
\item \textbf{Step 2:} Score the quality of each of the generated synthetic data points calculated by a \textbf{data value critic function/reward function}. We consider this score of a synthetic data point as \textbf{reward}.
\item \textbf{Step 3:} We use the output of the
data value critic function/reward function as a scalar reward. We fine-tune the generator to optimize this reward using \textbf{reinforcement learning}.
\end{itemize}
During the generation phase, the generated synthetic data is ranked by the reward function used in Step 2 so as to ensure the delivery of the best-quality samples to users. 

The overview of our method is shown in Figure~\ref{Fig:1}. The proposed method is called aligning \textbf{Syn}thetic data with human preferences using \textbf{R}einforcement \textbf{L}earning (\method). We will elaborate on the technical details in the following.

\subsection*{Generator model}
Our framework is model-agnostic: it applies to any differentiable synthetic data generator models (e.g. VAEs, GANs) because it is based on reinforcement learning. In this task, we pick a VAE generator model \texttt{TVAE} \cite{xu2019modeling} which has been of interest for generating tabular clinical trial data because of its ability to produce synthetic data that have relatively high fidelity to the source and can handle both categorical and continuous features.  Although we utilized \texttt{TVAE} in \method as the base generator model for the majority of experiments, other differentiable generative models can be used as the generator model.

\subsection*{Data value critic function/ Reward function}
One of the primary challenges that need to be resolved is the definition of \textit{data quality}, which should satisfy the following properties: (1) it should reflect the user's interests in a quantitative way; (2) it should be calculated in a cost and time efficient way to enable automatic updating of SDG models. However, evaluating the quality of synthetic patient data is considerably more complex than ranking the quality of generated texts through manual crowdsourcing used in RLHF for NLP tasks.

\noindent \textbf{Critic Function}. Our solution hinges on attaching a \textit{score} for each synthetic data based on a \textit{data value critic function}. The scores can then rank the quality of generated synthetic patient data as the feedback to improve the generator. To be specific, the data critic function is defined by
\begin{equation}\label{eq:critic_function}
    r_{\theta}(x_i, \hat x_i) = s_i - |x_i - \hat{x_i}|_{1}.
\end{equation}
Here, $s_i$ represents the \textit{utility} depending on the user's preferences; the second term indicates the \textit{fidelity} of the generated synthetic data $\hat{x}$ because it measures its distance to the real patient record $x$. It is noted that the $\ell_1$ norm is chosen over $\ell_2$ norm. The reason is that the $\ell_2$ norm tends to overweight continuous columns while underestimating the deviation of binary columns, which is not desirable for clinical trial data since many columns are binary, e.g., the event indicators. We used the \textit{fidelity} term as a regularizer to make sure that the model is not generating data with high utility and low fidelity, as both need to be high for synthetic clinical trial data.

\noindent \textbf{Utility Measurement}. One central challenge is quantifying the utility, denoted as $s_i$, based on user preferences. In this particular scenario, users will utilize the generated synthetic data for predictive modeling in a downstream task. For example, they may use the data to predict the incidence rate of a specific clinical endpoint for a patient, using the patient's records as input. In essence, the goal is to \textit{identify the synthetic patient data that maximizes the performance of the downstream task predictions}.

Denote the generated synthetic patient dataset by $\mathbf{I}=\{\hat{x}_1,\dots,\hat{x}_N\}$, which is used as the training data in the downstream task. From each patient record, we extract the specified clinical endpoint $y$ as the target label to predict, as $\{\hat{y}_1,\dots,\hat{y}_N\}$. Similarly, we denote the real patient data by $\mathbf{T}=\{x_1,\dots,x_M\}$ and the corresponding labels $\{y_1,\dots,y_M\}$. The predictive model, which could be any black-box algorithm, is fitted to predict $y$ given the input $x$, and then evaluated on the test set $\mathbf{T}$. We hence denote the performance score obtained by the model on the test set by $V(\mathbf{I},\mathbf{T})$ (e.g., accuracy, ROC-AUC, etc.), which measures to what extent the utility provided by the synthetic data to the downstream task. We omit $\mathbf{T}$ in $V(\cdot)$ from now on because it is always fixed. The object of interest here is to maximize $V(\mathbf{I})$ in order to provide high-quality synthetic data that satisfies user preferences.

\noindent \textbf{Data Shapley}. The $s_i$ in Eq. \eqref{eq:critic_function} works as a metric to assess the individual contribution of each synthetic data point to the value function $V$. Specifically, we employ data Shapley, a well-established method widely used for measuring the impact of a single data point on a specific target task \cite{ghorbani2019data}, defined by
\begin{equation}\label{eq:data_shapley}
    s_i=\frac{1}{N}\sum_{S \subseteq \mathbf{I}\setminus{\{\hat{x}_i\}}}\frac{1}{\binom{N-1}{|\mathbf{S}|}} [V(\mathbf{S} \cup \{\hat{x}_i\})-V(\mathbf{S})],
\end{equation}
where $\mathbf{S}$ is a subset of training dataset $\mathbf{I}$ excluding $\hat{x}_i$; $|\cdot|$ measures the cardinality of the set. Computing the exact value of data Shapley via Eq. \eqref{eq:data_shapley} requires an exponentially large number of computations with respect to the number of training data. Instead, we use K-Nerest-Neighbor data Shapley (KNN Shapley) \cite{jia2019efficient} as an approximation (See Appendix \ref{apd:first}). This algorithm works for $N$ training samples in $\mathcal{O}(N\log N)$ time.

\subsection*{Reinforcement learning}
The synthetic data generated by the generator model goes through the data value critic function to assign a reward $r_\theta(\cdot)$ to each synthetic data. These rewards can then be leveraged as the \textit{critic} produces real-time feedback, e.g., ranking of the generated synthetic records, for the generator model. We execute the RL process using the Proximal Policy Optimization (PPO) algorithm \cite{schulman2017proximal} to provide this feedback to the generator model. The setting is a bandit environment where a real clinical trial record $x$ is presented, and the system expects a corresponding synthetic clinical trial record $\hat{x}$ to be generated based on the actual record. Given the real clinical trial record and the synthetic record, it produces a reward determined by the reward function and ends the episode. We may repeat this process to obtain a set of synthetic records $\{\hat{x}_*\}$ based on $x$ for a more comprehensive ranking. We will then rank the data grounded on the data quality critic in Eq. \eqref{eq:critic_function}.
 
To prevent the model from being excessively optimized, we introduced a \textit{per-record penalty} that reduces the variance by controlling the distance between the synthetic records after and before the policy update, which renders the final objective function that we want to maximize:

\begin{equation}
    \ \mathcal{L}(\phi) = \bmE_{(x,\hat x) \sim D_{\pi^{\text{RL}}_{\phi}}} \left[r_{\theta}(x, \hat x) - \log (\frac{\pi^{\text{RL}}_{\phi}(\hat x|x)}{\pi^{\text{ref}}_{\phi}(\hat x^\prime|x)}) \right],
\end{equation}

where $\pi^{\text{RL}}_{\phi}(\hat x|x)$ is the learned RL policy,  $\pi^{\text{ref}}_{\phi}(\hat x^\prime|x)$ is the reference policy from the generator model before using RL. $x$ denotes one real patient data; $\hat{x}$ is one synthetic patient data generated based on the real patient data input $x$. The generator is then optimized to maximize the generated data quality adapted to the user preferences while maintaining fidelity.

\section{Experimental setup}

\subsection{Data Sources}
\noindent \textbf{Melanoma:} This dataset (NCT00522834) represents a Phase 3 trial evaluating the effectiveness of Elesclomol (STA-4783) when used in conjunction with Paclitaxel versus Paclitaxel as a standalone treatment for individuals with stage IV metastatic melanoma who have not undergone prior chemotherapy. Among the 651 participants initially involved, records for 326 patients have been made accessible to the public at Project Data Sphere \footref{first_footnote}. Following the processes of data cleaning and preprocessing, we now have a total of 310 patient records available for analysis.

\noindent \textbf{Breast Cancer:}  This dataset is a phase III breast cancer clinical trial (NCT00174655). There are a total of 2,887 patients who were randomly assigned to the arms to evaluate the activity of Docetaxel, given either sequentially or in combination with Doxorubicin, followed by CMF, in comparison to Doxorubicin alone or in combination with Cyclophosphamide, followed by CMF, in the adjuvant treatment of node-positive breast cancer patients. We downloaded and processed the publicly available dataset from Project Data Sphere \footnote{{\url{https://www.projectdatasphere.org}}\label{first_footnote}} that contains a total of 994 control participants. 

\noindent \textbf{NSCLC:} This dataset pertains to the clinical trial for Non-small cell lung cancer (NSCLC) registered under NCT00981058. Out of a total of 1093 trial participants, data from 548 patients have been made publicly accessible at Project Data Sphere \footref{first_footnote} for analysis. The primary aim of this investigation was to assess the overall survival of individuals diagnosed with Stage IV squamous NSCLC who underwent treatment with necitumumab in combination with gemcitabine and cisplatin chemotherapy, as compared to those who received gemcitabine and cisplatin chemotherapy alone.

\noindent \textbf{CAR-T:}  The fourth dataset is a collection of 27 clinical trials where CAR-T cell therapy (CAR-T and T-Cell Engager) has been administered. The CAR-T dataset includes 5,619 patients diagnosed with one of the following disease indications: (i) acute lymphoblastic leukemia, (ii) solid tumor, (iii) Non-Hodgkin's lymphoma, and (iv) B-lymphoblastic leukemia. All included patients were enrolled in the treatment arm. 

Details about data preprocessing and other implementation details are described in the Appendix \ref{apd:first}.

\subsection{Baseline algorithms}

The Synthetic Data Vault Python framework (version 1.0.0) was used to generate synthetic datasets to the real clinical trial datasets for the baseline algorithms (\texttt{CTGAN}, \texttt{CopulaGAN}, and \texttt{TVAE}) \cite{patki2016synthetic}. \method performance was compared against these baseline synthetic data generation algorithms.

\begin{itemize}[leftmargin=*]
    \item \texttt{CTGAN} is a conditional generative adversarial network designed to model the complex joint distribution of mixed-type tabular data \cite{xu2019modeling}.
    \item \texttt{CopulaGAN} is a variation of the \texttt{CTGAN} Model which utilizes the cumulative distribution function-based transformation through the \texttt{GaussianCopula} model \cite{sun2019learning, patki2016synthetic}.
    \item \texttt{TVAE} has the similar purpose as \texttt{CTGAN} that can handle mixed-type tabular data \cite{xu2019modeling}. 
\end{itemize}

\subsection{Evaluation Metrics}
We evaluate \method with the utility metric \textbf{ML Efficiency}, which provides an assessment of how well a synthetic clinical outcome or endpoint of interest tracks the properties of the source data.  We additionally assess fidelity and privacy, which are critical metrics for assessing synthetic clinical trial data \cite{beigi2022synthetic}. In this subsection, we describe briefly all the evaluation metrics we used for this study.\\

\noindent \textbf{Utility Metrics:} Synthetic data utility was measured by using \textbf{ML Efficiency}: the performance of the machine learning models used to predict scenarios of interest (e.g., clinical endpoints like death prediction, adverse event prediction). A Random Forest Classification model \cite{randomforest,sklearn_api} was trained to predict each outcome of interest (e.g., death flag and adverse event flag) using all other features as input. 10-fold cross-validation was performed where test data (10\% of the total dataset) were sampled from the real data where 90\% of the real data or the entire synthetic data was used as training data for the prediction model. We used the following ML Efficiency utility metrics:
\begin{itemize}[leftmargin=*]
\item \textbf{Area Under the Receiver Operating Characteristic curve (AUROC).} We utilized scikit-learn \cite{scikit-learn} to measure AUROC for binary classification scenarios.
\item \textbf{Mean Squared Error (MSE).} This measures the average of the squared errors or residuals between the predicted and true values. We utilized scikit-learn \cite{scikit-learn} to measure MSE for regression scenarios.
\end{itemize}

\noindent \textbf{\textit{Melanoma.}} ML efficiency was defined as the ability to predict the respiratory rate at the last visit of the patient. The performance of the prediction model was measured using MSE where training data came from different synthetic data generation models. \\

\noindent \textbf{\textit{Breast Cancer.}} ML efficiency was defined as the ability to predict occurrence of an adverse event of interest. The performance of the prediction model was measured using AUROC where training data came from different synthetic data generation models.\\ 

\noindent \textbf{\textit{NSCLC.}} ML efficiency was defined as the ability to predict the best response of the patient. There were 6 unique best responses (from 0 to 5). We converted the problem to a binary classification task by splitting the classes into two bigger classes. Class 0: best responses 0 to 3, Class 1: best responses 4 to 5. \\

\noindent \textbf{\textit{CAR-T.}} ML efficiency for this data was centered around predicting the binary death flag. The performance of the prediction model was measured using AUROC for death flag where training data came from different synthetic data generation models.\\

\noindent \textbf{Fidelity Metrics:} We used the following evaluation metrics for accessing fidelity of the synthetic data:
\begin{itemize}[leftmargin=*]
    \item \textbf{Silhouette coefficient}: This metric indicates the goodness of clustering \cite{rousseeuw1987silhouettes} where values closer to 1 indicate separable clusters and values closer to 0 indicate overlapping clusters. Given labels for synthetic and real data, the silhouette coefficient close to 0 indicates samples could belong either to synthetic or real data whereas higher values indicate separability. This score was computed using the scikit-learn Python library (version 1.0.2) \cite{sklearn_api}.
    
    \item \textbf{Percentage of dissimilar columns.} This metric is defined as the percentage of columns in the synthetic dataset where the distribution of values was determined as significantly different using the Kolmogrov-Smirinov test \cite{massey1951kolmogorov} (using a statistical significance p-value threshold $\alpha < 0.05$). Values close to 0 indicate a lower number of dissimilar columns between synthetic and real data and a higher quality of synthetic data.
    
    \item \textbf{Column shapes.} This metric quantifies the average similarity between real and synthetic data by assessing the resemblance of numerical columns using the Kolmogorov-Smirnov test \cite{massey1951kolmogorov} and categorical columns by comparing value frequencies. It is calculated as the mean similarity across all columns in the synthetic data, with values approaching 1 indicating strong similarity. This metric is sourced from Python SDV library \cite{patki2016synthetic}.

    \item \textbf{Column pair trends.} This metric indicates the similarity in pairwise correlations across columns between real vs. synthetic data. This metric uses Pearson correlation and total variation distance for numerical and categorical columns respectively \cite{freedman2007statistics}. Values close to 1 indicate high similarity in pairwise correlations in synthetic compared to real data. This metric is sourced from Python Synthetic Data Vault library \cite{patki2016synthetic}.
    
\end{itemize}

\noindent \textbf{Privacy Metrics:} We used the following metrics to show the extent to which the synthetic data protects the privacy of individuals in real clinical trial data:
\begin{itemize}[leftmargin=*]
    \item \textbf{Mean inference risk.} This test assesses the mean inference risk across all columns in the synthetic data where the inference risk for each column is computed as the attacker's ability to predict the attribute for an original record using the synthetic data.  Lower values close to 0 indicate a lower risk of disclosure using this attack. Python library \texttt{anonymeter} \cite{anonymeter} was used to evaluate this metric.
    \item \textbf{Privacy loss.} This metric is calculated as the difference between the adversarial test accuracy and adversarial train accuracy \cite{yale2019privacy}. Values closer to 0 denote low privacy loss whereas high values close to 1 indicate higher loss of privacy. 
\end{itemize}

\section{Results} \label{sec:experiment}
In this section, we present the experiment results that evaluate \method across four clinical trial datasets in three aspects: (i) \textbf{Utility}: the ability of synthetic data to reproduce the downstream tasks of interest, (ii) \textbf{Fidelity}: the degree to which the synthetic data generated by \method resembles real clinical trial data, and (iii) \textbf{Privacy}: the extent to which the synthetic data protects the privacy of individuals in the real trial data. We also include results for \method's application as a general framework where \texttt{CTGAN} replaces \texttt{TVAE} as the base generator model.

\subsection{Utility evaluation}
Synthetic data utility was defined as the performance of the machine learning model in predicting the downstream task of interest (or \textbf{ML Efficiency}) when trained on the synthetic training data. Any improvements of \textbf{ML Efficiency} for datasets generated using \method compared to the base model \texttt{TVAE} would indicate improved utility of \method. Utility evaluation results compared \method to \texttt{TVAE} across four clinical datasets as shown in Table \ref{tab: utility}. For Breast Cancer, CAR-T and NSCLC datasets categorical variables were selected as the downstream task to be predicted (Adverse event of interest, Death status of patient, Best overall response of patient respectively) and \textit{AUROC} metric was used to evaluate these datasets (Ideal score: 1). The objective for the task on the Melanoma dataset was to predict respiratory rate (a continuous numeric outcome), where \textit{MSE} was the evaluation metric (Ideal score: 0). Across all datasets, \method outperformed its base model \texttt{TVAE}, improving utility \textbf{ML Efficiency} scores with lowered \textit{MSE} for Melanoma dataset task, and increased \textit{AUCROC} scores on the tasks for Breast Cancer, NSCLC, and CAR-T (Table \ref{tab: utility}). Additionally, \method improved over the utility scores for other base generator models \texttt{CTGAN} and \texttt{CopulaGAN}.

\subsection{Fidelity evaluation}
Fidelity of the synthetic clinical trial datasets was evaluated to assess the similarity of synthetic data to the real clinical trial data. Metrics used to evaluate fidelity were: (i) \textit{Silhouette coefficient} \cite{rousseeuw1987silhouettes} (Ideal score: 0), (ii) \textit{\% Dissimilar columns} (Ideal score: 0), (iii) \textit{Column shapes}\cite{patki2016synthetic} (Ideal score: 1), and (vi) \textit{Column pair trends}\cite{patki2016synthetic} (Ideal score: 1). Fidelity of synthetic datasets produced using \method was compared to that of datasets produced using the base generator model \texttt{TVAE}. Table \ref{tab: fidelity} shows that fidelity scores for \method were comparable to \texttt{TVAE} across the four clinical datasets. Results comparing \method to all baseline algorithms (\texttt{CTGAN}, \texttt{CopulaGAN}, \texttt{TVAE}) are included in Appendix \ref{apd:first}.

\subsection{Privacy Evaluation}
\begin{figure}[t]
  \centering
  \includegraphics[width=.9\linewidth]{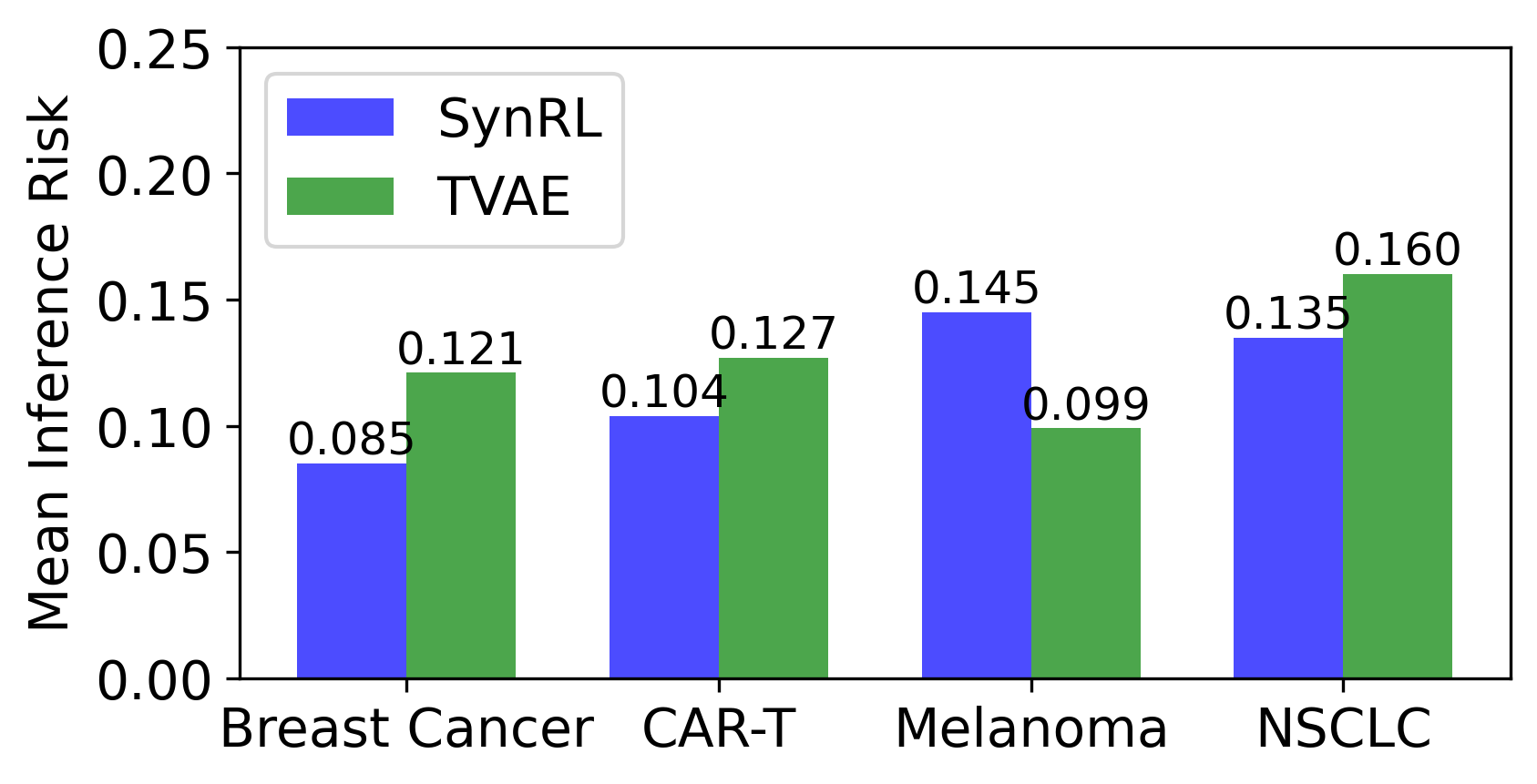}
  \caption{Mean Inference Risk (lower better) comparison between \texttt{TVAE} and \method.}
  \label{fig: privacy1}
\end{figure}

\begin{figure}[t]
  \centering
  \includegraphics[width=.9\linewidth]{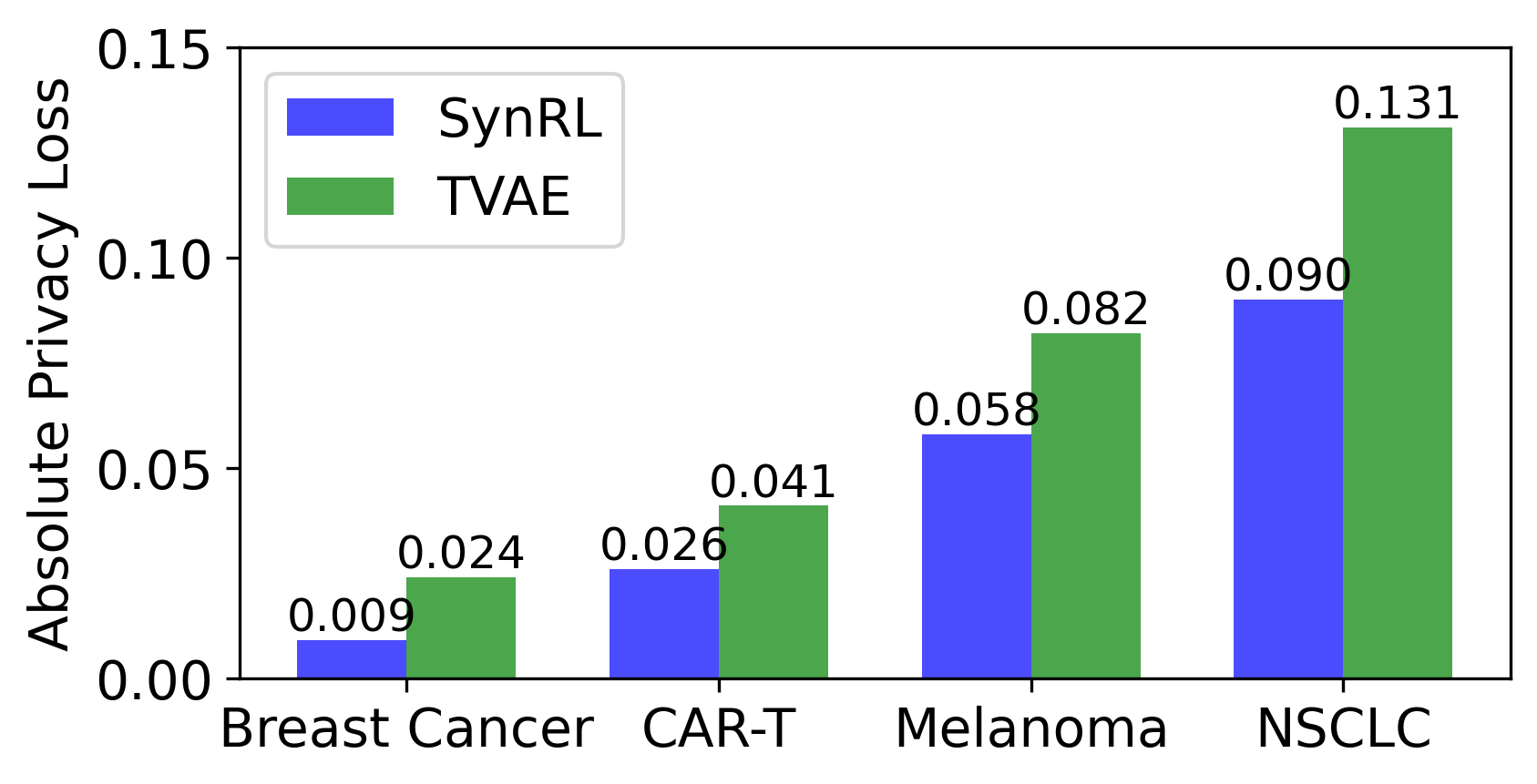}
  \caption{Absolute Privacy Loss (lower better) comparison between \texttt{TVAE} and \method.}
  \label{fig: privacy2}
\end{figure}

Privacy preservation provided by synthetic data generation algorithms was evaluated across four clinical datasets using the following metrics: (i) \textit{mean inference risk} \cite{anonymeter} (Ideal value: 0) and (ii) \textit{privacy loss}  \cite{yale2019privacy} (Ideal value: 0). The data was divided using an 80-20 training and testing data split where the training data was used to train the synthetic data generator models and 20\% data for retained for testing. An equivalent volume of synthetic data was produced by each synthetic data generation algorithm that matched the size of the training data. Figures \ref{fig: privacy1} and \ref{fig: privacy2} show the privacy evaluation scores for \method compared to \texttt{TVAE} across all four datasets. Synthetic datasets generated using \method had lower \textit{privacy loss} scores when compared to \texttt{TVAE} across all clinical datasets. Lower \textit{mean inference risk} for across 3 out of 4 clinical datasets (except Melanoma dataset) was also observed for synthetic datasets generated using \method. Privacy results comparing \method performance to all other baselines are included in Appendix \ref{apd:first}.

\begin{table*}[!htb]
\centering
\caption{ML efficiency results from 10-fold cross-validation.  \texttt{Real} denotes the performance of the prediction model where only real training data was used for training and performance was tested on real test data. For all other rows, synthetic data generated using the respective generative model was used to train the prediction model and tested on real test data.  Best performance is highlighted in \textbf{bold} per column (among baseline models).}
\label{tab: utility}
\resizebox{\textwidth}{!}{%
\begin{tabular}{@{}lllll@{}}
\toprule
          & Melanoma   & Breast Cancer                                 & NSCLC    & CAR-T                \\ \midrule
          & MSE (lower better)                 & AUROC (higher better)  & AUROC (higher better)                                    & AUROC (higher better)                   \\ \midrule
Real              & 4.47$\pm$ 0.73        & 0.87 $\pm$ 0.18              & 0.88 $\pm$ 0.04    & 0.65 $\pm$ 0.06     \\ \midrule
\texttt{CTGAN}      & 5.42 $\pm$ 0.61        & 0.85 $\pm$ 0.17                 & 0.49 $\pm$ 0.08      & 0.51 $\pm$ 0.07      \\
\texttt{CopulaGAN}    & 6.28 $\pm$ 1.24       & 0.85 $\pm$ 0.18                & 0.38 $\pm$ 0.08         & 0.51 $\pm$ 0.05  \\
\texttt{TVAE}          & 6.25 $\pm$ 1.18        & 0.89 $\pm$ 0.16        & 0.54 $\pm$ 0.07       & 0.65 $\pm$ 0.05  \\
\method    & \textbf{4.99 $\pm$ 0.70}   & \textbf{0.91 $\pm$ 0.15} & \textbf{0.58 $\pm$ 0.09}  & \textbf{0.67 $\pm$ 0.06}\\ \bottomrule
\end{tabular}%
}
\end{table*}

\begin{table*}[t]
\centering
\caption{Fidelity Scores }
\label{tab: fidelity}
\resizebox{\textwidth}{!}{%
\begin{tabular}{@{}|l|ll|ll|ll|ll|@{}}
\toprule
                    & \multicolumn{1}{l}{Melanoma} & \multicolumn{1}{l|}{}  & \multicolumn{1}{l}{Breast Cancer} & \multicolumn{1}{l|}{}                  & \multicolumn{1}{l}{NSCLC} & \multicolumn{1}{l|}{}    & \multicolumn{1}{l}{CAR-T} & \multicolumn{1}{l|}{}  \\ \midrule
                     & \multicolumn{1}{l}{TVAE}          & \multicolumn{1}{l|}{\method} & \multicolumn{1}{l}{TVAE}  & \multicolumn{1}{l|}{\method} & \multicolumn{1}{l}{TVAE}     & \multicolumn{1}{l|}{\method} & \multicolumn{1}{l}{TVAE}  & \multicolumn{1}{l|}{\method} \\ \midrule
Silhouette score     & 0.082                        & 0.085  & 0.014               & 0.013                                                                 & 0.070                      & 0.070      & 0.055                     & 0.048                 \\
\% Dissimilar columns  & 0.456                        & 0.491 & 0.911                             & 0.893                                                                 & 0.317                     & 0.317             & 0.911                     & 0.885          \\
Column Shapes       & 0.875       & 0.872  & 0.901            & 0.898                                                        & 0.912       & 0.912         & 0.935         & 0.936            \\
Column Pair Trends   & 0.936         & 0.941 & 0.831           & 0.822                                                     & 0.951                     & 0.948           & 0.889          & 0.890           \\ \bottomrule
\end{tabular}%
}
\end{table*}

\noindent
\subsection{\method as a general framework} \label{sec:general}
\begin{figure}[h]
  \centering
  \includegraphics[width=0.5\textwidth]{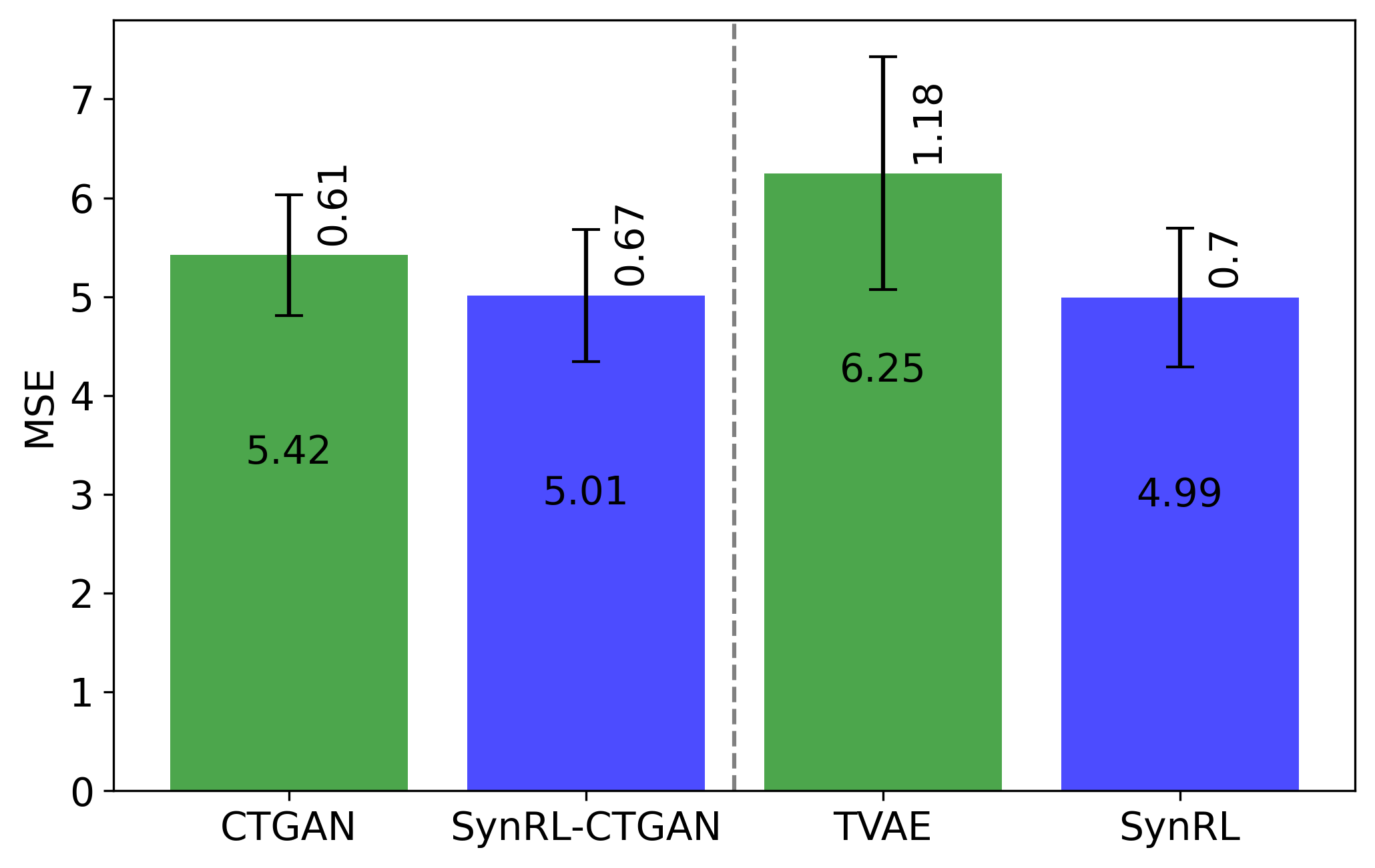}
  \caption{\method as a general framework for customizing synthetic data from multiple base generators.}
  \label{fig:3}
\end{figure}

\method can be utilized as a general framework where other deep learning generative methods can substitute \texttt{TVAE} as the base generator of choice. To show the use of \method as a general framework, we replaced the base model with \texttt{CTGAN} and evaluated the performance of the models \texttt{SynRL-CTGAN} and \texttt{CTGAN} on the Melanoma dataset (Figure \ref{fig:3}). Figure \ref{fig:3} shows the utility evaluation results on the Melanoma dataset for the synthetic data generation models: \texttt{SynRL-CTGAN}, \texttt{CTGAN}, \texttt{TVAE}, and \method (\method as a framework using \texttt{TVAE} as a base model). When comparing utility, \texttt{SynRL-CTGAN} performed better than \texttt{CTGAN} in the downstream prediction task for the Melanoma dataset (predicting the respiratory rate of the last visit of patients). \method as a framework showed consistent improvement in utility over the base model (\texttt{TVAE} and \texttt{CTGAN}) regardless of the selected base model.

\subsection{Comparison of \method with modified optimization function} \label{predictionhead}
We have experimented with adding prediction loss directly to the base model's optimization function. Table \ref{tab:Comparison_pred_head } shows that \texttt{SynRL} and \texttt{TVAE} performs better in terms of utility and fidelity compared to adding a prediction head to \texttt{TVAE}. In contrast, \method utilizes the reward function (balancing fidelity and utility terms) where the RL approach ensures the synthetic data generated by the pre-trained base generator does not change drastically from the real data. We performed additional experiments on the Melanoma dataset to compare the impact of prediction loss versus RL by adding prediction head to \texttt{TVAE} with a prediction loss. The results (Table \ref{tab:Comparison_pred_head }) indicate with increasing importance of prediction loss using hyperparameter $\alpha$ utility improves (decreased MSE values), however fidelity decreases. Though the utility scores for TVAE+prediction loss improve with increasing alpha, the utility scores for TVAE and \method still outperform the best \texttt{TVAE} + prediction loss performance.Therefore, we posit that \method outperforms \texttt{TVAE} with multiple losses (VAE loss + prediction loss) trained on a smaller clinical trial dataset. For a fair comparison between \method and \texttt{TVAE} + prediction loss, we optimized the \texttt{TVAE} + prediction loss model by varying the number of layers and size of the layers of the prediction head. Table \ref{tab:Comparison_pred_head } only shows the results for the best \texttt{TVAE} + prediction loss model.
\begin{table*}[]
\centering
\begin{tabular}{@{}llll@{}}
\toprule
\textbf{}                   & \textbf{Utility} & \textbf{Fidelity} & \textbf{}          \\ \midrule
                            & MSE              & Column Shapes     & Column pair trends \\
\method                       & 4.99±0.70        & 0.872             & 0.941              \\
\texttt{TVAE}                        & 6.25±1.18        & 0.875             & 0.936              \\
\texttt{TVAE} + 1* prediction loss   & 64.08 ± 5.18     & 0.857             & 0.577              \\
\texttt{TVAE} + 100* prediction loss & 44.60 ± 4.22     & 0.859             & 0.549              \\
\texttt{TVAE} + 500* prediction loss & 44.41± 4.21      & 0.846             & 0.514              \\ \bottomrule
\end{tabular}
\caption{\texttt{TVAE} and \texttt{TVAE} + $\alpha$ * prediction loss both were trained for 500 epochs. Pretrained \texttt{TVAE} was used as base generator in \texttt{SynRL}. Here alpha is a hyperparameter whose values are selected from {[}1, 100, 500{]}. The prediction head has 4 linear layers with the sizes of {[}128,128, 64, 32{]}.}
\label{tab:Comparison_pred_head }
\end{table*}

\section{Discussion} \label{sec: discussion}
Though synthetic data generation methods like VAEs and GANs have shown the ability to generate synthetic clinical trial data while preserving the privacy and fidelity, the ability to reproduce valuable downstream analysis remains limited (\cite{beigi2022synthetic}). The proposed \method  overcomes this limitation by using the reward function and the RL feedback loop to improve the utility of the synthetic dataset on the downstream task of interest. Across clinical trial datasets of varying sizes and downstream tasks of categorical and numerical outcome predictions, \method outperformed all methods including its own base model (\texttt{TVAE}) in improving utility (Table \ref{tab: utility}) while preserving fidelity and privacy scores similar to the base model(Table \ref{tab: fidelity}, Figure \ref{fig: privacy1} and Figure \ref{fig: privacy2}). Though performance results for other synthetic data generation models are included for overall comparison, \method fulfills the main criteria of success in improving performance of the base generator model for downstream prediction tasks. In addition to utility, a critical use case of synthetic data generation in healthcare is the privacy protection it provides to patient data. \method performed better than \texttt{TVAE} in all cases except for the Melanoma dataset for mean inference risk. As this metric is average inference risk over all attributes, it can be influenced by outliers. The results also support that the \method framework is capable of handling variational autoencoder-based and GAN-based base generator models to finetune and customize the synthetic data generation for clinical trials. The Melanoma dataset (n = 326) is the smallest dataset (NSCLC(n=548), Breast cancer(n=994), CAR-T(n=5,619)) in our set of experiments. Given the typically small sample sizes in clinical trials, which can range from tens to a few thousands, we chose the Melanoma dataset to demonstrate \method's performance on a real-world clinical trial dataset, showcasing the generalizability of \method as discussed in Section \ref{sec:general} and showing \method's comparison with \texttt{TVAE}+prediction loss in Section \ref{predictionhead} . It is important to note that alternative methods such as adding prediction loss to the base model's optimization function can be used to generate synthetic data sets that are optimized on downstream prediction tasks of interest. However, these methods may suffer in balancing utility versus fidelity resulting in compromised fidelity while optimizing over utility (Appendix \ref{predictionhead}).

\method demonstrates its adaptability across various deep learning generative methods as base generators, as evidenced by its enhanced performance on real-world clinical trial data. This flexibility underscores its efficacy in tailoring synthetic data generation for diverse healthcare contexts, showcasing its potential as a versatile tool for advancing machine learning applications beyond clinical trial data synthesis.

\subsection{Limitations} \label{sec: limitation}
A limitation of our work is that the user preference is limited to a downstream task where any constraints for the synthetic data generation (e.g., any constraint on number of medications provided per visit) can not be handled by the current \method framework. However, as long as the constraint is quantifiable, we can add that to the reward model for constrained generation. We would like to address this limitation in our future work. \method is limited to tabular datasets where further work is needed to synthesize sequential data.

\section{Conclusion} \label{sec: conclusion}
We have introduced a framework for refining synthetic clinical trial data generator models for users' preferred clinical endpoints and outcomes of interest using a reward model and reinforcement learning. It is crucial to emphasize that the primary aim of this approach is to create synthetic data that enhances utility metrics while preserving the fidelity and privacy of the original generation model. The results indicate that \method generates synthetic clinical trial data that outperforms those generated through the base generator model when considering the utility of the prediction tasks of interest. Moreover, \method fidelity outcomes are comparable to those of the base generator method and \method privacy results are better than the base generator method, implying that there are minimal data-sharing concerns added by the framework. Though \texttt{TVAE} was used as the underlying generator model in \method for the current set of experiments, the proposed framework of fine-tuning synthetic data generation using a reward model and reinforcement learning is agnostic to \texttt{TVAE} and can be replaced with other generators. Moreover, further investigation is needed to evaluate the feasibility of surrogate utility measurements in aligning synthetic data generation. Additionally, we also aim to incorporate a privacy component into the reward function to enhance privacy. Overall, \method provides a general framework for increasing the value of synthetic clinical trial data within the many applications that require high utility of specific clinical outcomes and endpoints.



\section{Conflicts of interest}
The authors have no competing interests to declare.

\bibliography{custom}
\bibliographystyle{acl_natbib}

\newpage
\appendix

\section{}\label{apd:first}

\subsection{Data preprocessing} 

\noindent \textbf{Melanoma.}
This data comprised of baseline patient information, treatments, medications, and adverse events experienced by patients during the trial. From a cohort of 651 patients in the clinical trial, only 326 patients' data were publicly available, where 16 patients were excluded due to high missingness across columns, resulting in a final cohort of 310 patients. We used 3 baseline features (i.e., age, sex, and race) and frequencies of all the other features that occurred over time for each patient.

\noindent \textbf{Breast Cancer.} This data comprised baseline patient information and adverse events experienced by patients during the trial. From a cohort of 2,887 patients in the clinical trial, only 994 patients' data were publicly available, where 25 patients were excluded due to high missingness across columns, resulting in a final cohort of 969 patients. Of the total of 68 features in the dataset, 11 features represented baseline information (i.e., height, weight, age, race, number of relapse, tumor size, multi-focal tumor presence, number of positive axillary lymph nodes, tumor location, histopathologic grade, and histopathologic type) and 57 binary features represented adverse events experienced by the patients (indicating presence or absence of the specific adverse event). We discarded medications and adverse events that are present less than 200 times in total 8293 visits of all patients.

\noindent\textbf{NSCLC.}
This data comprised baseline patient information, medications, and adverse events experienced by patients during the trial. From a cohort of 1093 patients in the clinical trial, only 548 patients' data were publicly available, where 21 patients were excluded due to high missingness across columns, resulting in a final cohort of 527 patients. We used 3 baseline features (i.e., age, sex, and race) and frequencies of all the other features that occurred over time for each patient.

\noindent \textbf{CAR-T.} This data comprised of baseline patient information and adverse events experienced by patients during the trial. Of the 5,619 patients in the cohort, we filtered out any patients where death happened 1000 days after treatment, resulting in 4,683 patients in the final cohort. Of the total of 164 features in the data, 12 features represented baseline information (i.e., region, sex, enrollment flag, randomization flag, end of study status, end of treatment status, death flag, age, race, disease indication, cancer stage at screening, patient status at the end of CAR-T treatment) and 152 binary features represented adverse events experienced by the patients (indicating presence or absence of the specific adverse event). From the set of total adverse events available, all rare adverse events that were present in less than 5\% of the total cohort were removed, resulting in the 152 adverse events in the tabular data used for training.

\subsection{Implementation Details}
We used PyTorch (\cite{NEURIPS2019_9015}) for our implementation. For the base generator model, we used TVAE's and CTGAN's implementation from \href{https://github.com/sdv-dev/CTGAN}{https://github.com/sdv-dev/CTGAN}. For ML efficiency, we used Random forest (RF) classifiers from \href{https://scikit-learn.org/stable/}{https://scikit-learn.org/stable/}.  We train the base generator TVAE for 400 epochs for both Breast cancer and CAR-T datasets. For RL feedback, we use 50 updating epochs to update the base generator model for the CAR-T dataset and 2 updating epochs for the breast cancer dataset.  We train the base generator TVAE for 500 epochs for both melanoma and NSCLC datasets. For RL feedback, we use 10 updating epochs to update the base generator model for the NSCLC dataset and 200 updating epochs for the melanoma dataset. Users can specify how many times they want to use RL feedback loop according to their need for their specific datasets. We stopped feedback epochs by looking at the loss curve and stopped when the loss seemed not decreasing. KNN-Shapley values calculation for synthetic data scales with O(NlogN) complexity, influenced by the number of epochs used for fine-tuning. For instance, SynRL takes 2-3 mins to fine-tune on a Melanoma dataset (N=326 patients) for 10 epochs with an NVIDIA GeForce RTX 2080 Ti GPU. While typical clinical trial datasets have similar run-times, larger datasets may take longer, where mini-batch techniques can aid in reducing computational complexity. For smaller datasets like the Melanoma dataset ( n = 326), conducting KNN-Shapley calculations on the entire dataset is recommended due to potential fragmentation of nearest neighbors in mini-batches. Real-world clinical trial datasets, typically small, further support this approach. However, for larger datasets, employing a mini-batch method can offer faster processing, albeit with potential quality trade-offs in nearest neighbor identification.

\begin{figure}[!htb]
  \centering
 \includegraphics[width=0.7\linewidth]{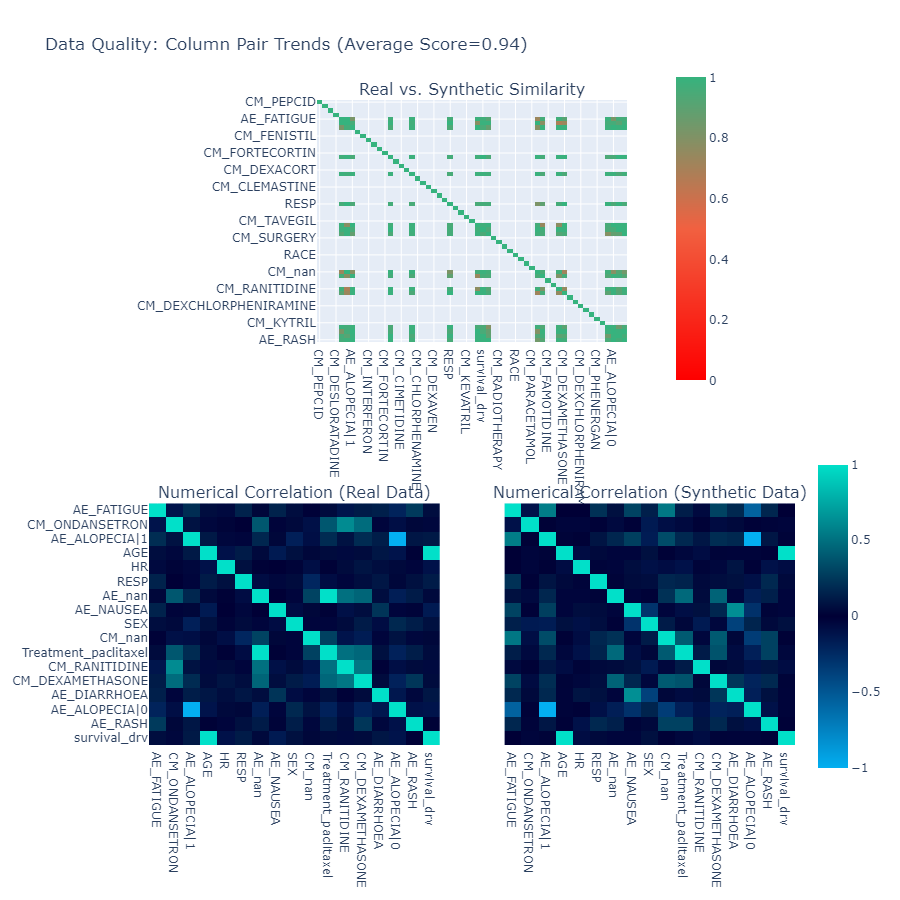}
  \caption{Column pair trends of synthetic data generated by \method (Melanoma) }
  \label{fig:4}
\end{figure}

\begin{table}[!htb]
\centering
\caption{Fidelity scores for different datasets (Silhouette score and Column pair trends).}
\label{tab:fidelity_combined_1}
\resizebox{\columnwidth}{!}{%
\begin{tabular}{@{}lrrrrrrrr@{}}
\toprule
\textbf{Dataset} & \multicolumn{4}{c}{\textbf{Silhouette score}} & \multicolumn{4}{c}{\textbf{Column pair trends}} \\
\cmidrule(lr){2-5} \cmidrule(lr){6-9}
 & \texttt{CTGAN} & \texttt{CopulaGAN} & \texttt{TVAE} & \texttt{SynRL} & \texttt{CTGAN} & \texttt{CopulaGAN} & \texttt{TVAE} & \texttt{SynRL} \\
\midrule
Melanoma & 0.042 & 0.118 & 0.082 & 0.085 & 0.955 & 0.953 & 0.936 & 0.941 \\
Breast Cancer Data & 0.184 & 0.115 & 0.014 & 0.013 & 0.887 & 0.907 & 0.831 & 0.822 \\
NSCLC & 0.029 & 0.599 & 0.07 & 0.07 & 0.97 & 0.969 & 0.951 & 0.948 \\
CAR-T Data & 0.015 & 0.081 & 0.055 & 0.048 & 0.888 & 0.894 & 0.889 & 0.89 \\
\bottomrule
\end{tabular}%
}
\end{table}

\begin{table}[!htb]
\centering
\caption{Fidelity scores for different datasets (\% Dissimilar columns and Column Shapes).}
\label{tab:fidelity_combined_2}
\resizebox{\columnwidth}{!}{%
\begin{tabular}{@{}lrrrrrrrr@{}}
\toprule
\textbf{Dataset} & \multicolumn{4}{c}{\textbf{\% Dissimilar columns}} & \multicolumn{4}{c}{\textbf{Column Shapes}} \\
\cmidrule(lr){2-5} \cmidrule(lr){6-9}
 & \texttt{CTGAN} & \texttt{CopulaGAN} & \texttt{TVAE} & \texttt{SynRL} & \texttt{CTGAN} & \texttt{CopulaGAN} & \texttt{TVAE} & \texttt{SynRL} \\
\midrule
Melanoma & 29.825 & 49.123 & 45.614 & 49.123 & 0.921 & 0.809 & 0.875 & 0.872 \\
Breast Cancer Data & 34.056 & 34.783 & 91.129 & 89.344 & 0.927 & 0.945 & 0.901 & 0.898 \\
NSCLC & 24.038 & 36.538 & 31.731 & 31.731 & 0.946 & 0.766 & 0.912 & 0.912 \\
CAR-T Data & 73.837 & 80.117 & 91.096 & 88.513 & 0.925 & 0.931 & 0.935 & 0.936 \\
\bottomrule
\end{tabular}%
}
\end{table}

\begin{table}[!htb]
\centering
\caption{Privacy scores for different datasets.}
\label{tab:privacy_combined}
\resizebox{\columnwidth}{!}{%
\begin{tabular}{@{}lrrrrrrrr@{}}
\toprule
\textbf{Dataset} & \multicolumn{4}{c}{\textbf{Mean Inference risk}} & \multicolumn{4}{c}{\textbf{Privacy Loss}} \\
\cmidrule(lr){2-5} \cmidrule(lr){6-9}
 & \texttt{CTGAN} & \texttt{CopulaGAN} & \texttt{TVAE} & \texttt{SynRL} & \texttt{CTGAN} & \texttt{CopulaGAN} & \texttt{TVAE} & \texttt{SynRL} \\
\midrule
Melanoma & 0.098 & 0.092 & 0.099 & 0.145 & -0.062 & -0.195 & -0.082 & -0.058 \\
Breast Cancer Data & 0.112 & 0.1 & 0.121 & 0.0852 & -0.057 & -0.114 & -0.024 & -0.009 \\
NSCLC & 0.124 & 0.109 & 0.160 & 0.135 & -0.065 & 0.000 & -0.131 & -0.090 \\
CAR-T Data & 0.084 & 0.07 & 0.127 & 0.104 & -0.046 & -0.036 & -0.041 & -0.026 \\
\bottomrule
\end{tabular}%
}
\end{table}


\subsection {Additional Evaluation Results} \label{sec: additional}
\begin{figure*}[t]
  \centering
 \includegraphics[width=0.9\linewidth]{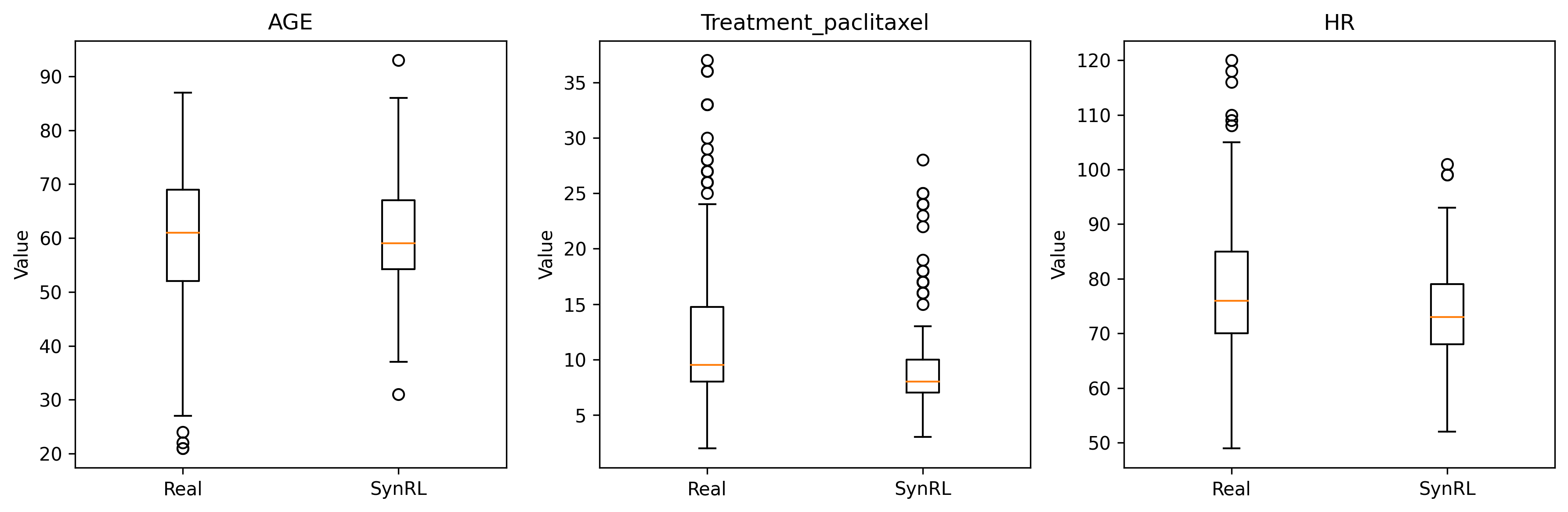}
  \caption{Boxplots for few example features from the Melanoma dataset.}
  \label{fig:7}
\end{figure*}

Tables  \ref{tab:fidelity_combined_1} and \ref{tab:fidelity_combined_2} show the fidelity results for all baseline models and \method. Figure \ref{fig:4}  shows the heatmaps indicating the correlation across features in the original and synthetic data for the Melanoma dataset. Each heatmap indicates the correlation across features where higher values in the heatmap indicate the two features in the dataset are highly correlated and vice versa. Similarity between the two heatmaps indicates that the synthetic dataset is able to preserve the correlations between features as observed in the original dataset \cite{patki2016synthetic}. As seen in Figure \ref{fig:4}, the heatmaps for the original and synthetic Melanoma datasets look similar indicating the preservation of the correlations across features. Figure \ref{fig:4} also shows the heatmap of “Synthetic vs. Real Similarity” of features where the similarity of each of the pairwise correlations between features in the synthetic dataset compared to the original dataset is reported. High similarity values indicate the pairwise correlation values for those features were similar in the synthetic data compared to the original data and vice versa. 
Table \ref{tab:privacy_combined} shows the privacy results for all baseline models and \method. Table \ref{tab:quality} shows few example rows generated by \method. Figure \ref{fig:7} shows boxplots for the distribution of selected features from the original and synthetic Melanoma dataset: Age, Number of times a patient received the treatment paclitaxel, and Heart rate. We can see the medians of the real feature values are similar to their corresponding synthetic ones (Figure \ref{fig:7}). Though the medians across the features were similar, the 25th and 75th quartile values for specific features (Number of times a patient received the treatment paclitaxel, and Heart rate) in the synthetic data were closer to the median compared to the original data.

\begin{table*}[h]
\centering
\caption{Example samples from synthetic Melanoma trial data. Columns starting with ‘CM\_’ represent medications, columns starting with ‘AE\_’ represent adverse events. Columns starting with ‘Treatment\_’ represent treatments being tested in this trial. HR represents heart rate.}
\label{tab:quality}
\resizebox{\textwidth}{!}{%
\begin{tabular}{|l|l|l|l|l|l|l|l|l|l|}
\hline
           & \textbf{AGE} & \textbf{SEX} & \textbf{RACE} & \textbf{Treatment\_paclitaxel} & \textbf{CM\_BENADRYL} & \textbf{…} & \textbf{AE\_FATIGUE} & \textbf{AE\_NAUSEA} & \textbf{HR} \\ \hline
\textbf{0} & 40           & 1            & 0             & 8                              & 0                     & …          & 1                    & 1                   & 74          \\ \hline
\textbf{1} & 38           & 1            & 0             & 8                              & 0                     & …          & 0                    & 0                   & 76          \\ \hline
\textbf{2} & 48           & 1            & 0             & 9                              & 0                     & …          & 0                    & 0                   & 76          \\ \hline
\textbf{3} & 66           & 0            & 0             & 10                             & 0                     & …          & 0                    & 0                   & 67          \\ \hline
\textbf{4} & 45           & 1            & 0             & 9                              & 0                     & …          & 0                    & 0                   & 78          \\ \hline
\textbf{5} & 73           & 1            & 0             & 10                             & 0                     & …          & 1                    & 1                   & 74          \\ \hline
\end{tabular}
}
\end{table*}

\end{document}